# Predicting electrode array impedance after one month from cochlear implantation surgery

Yousef A. Alohali
*College of Computer and Information Sciences*
*King Saud University*
*Riyadh, Saudi Arabia*
yousef@ksu.edu.sa

Yassin Abdelsamad
*Research Department*
*MED-EL GmbH*
*Riyadh, Saudi Arabia*
yassin.samad@gmail.com

Tamer Mesallam
*Research Chair of Voice, Swallowing, and Communication Disorders*
*King Saud University*
*Riyadh, Saudi Arabia*
tmesallam@ksu.edu.sa

Fida Almuhawas
*King Abdullah Ear Specialist Center*
*College of Medicine*
*King Saud University*
*Riyadh, Saudi Arabia*
fmuhawas@ksu.edu.sa

Abdulrahman Hagr
*King Abdullah Ear Specialist Center*
*College of Medicine*
*King Saud University*
*Riyadh, Saudi Arabia*
dhagr90@gmail.com

Mahmoud S. Fayed
*College of Computer and Information Sciences*
*King Saud University*
*Riyadh, Saudi Arabia*
msfayed@ksu.edu.sa

*Abstract*— Sensorineural hearing loss can be treated using Cochlear implantation. After this surgery using the electrode array impedance measurements, we can check the stability of the impedance value and the dynamic range. Deterioration of speech recognition scores could happen because of increased impedance values. Medicines used to do these measures many times during a year after the surgery. Predicting the electrode impedance could help in taking decisions to help the patient get better hearing. In this research we used a dataset of 80 patients of children who did cochlear implantation using MED-EL FLEX28 electrode array of 12 channels. We predicted the electrode impedance on each channel after 1 month from the surgery date. We used different machine learning algorithms like neural networks and decision trees. Our results indicates that the electrode impedance can be predicted, and the best algorithm is different based on the electrode channel. Also, the accuracy level varies between 66% and 100% based on the electrode channel when accepting an error range between 0 and 3 KO. Further research is required to predict the electrode impedance after three months, six months and one year.

## I. INTRODUCTION

Hearing loss is a known public health issue affecting older adults [1]. Also, children could suffer from it. Cochlear implantation is a surgical procedure which is performed for severely to profound hearing-impaired patients. Now in its fourth decade, Cochlear implants (CI) have successfully restored functional hearing to hundreds of thousands of individuals [2]. Cochlear implant system mainly includes audio processor, transmitter, receiver, and internal device which include the electrode array. Technically the external audio processor of CI collects the sounds through its microphone, then converts it into detailed digital signals. Therefore, and using speech and signal processing algorithms, this is signal is transmitted to the internal device via an inductive link. The internal device then converts this signal to electric impulses and pass them to the inner ear through the intracochlear electrode array [3]. CI companies offer different devices which vary in electrode size, shape, number of contacts, and configuration [4]. There is a positive correlation between the daily duration of the audio processor usage and speech performance. For pediatric patients, there is a special need to use implanted device for at least 8.3 hours/day to achieve acceptable language development [5]. CI recipients could reach and acceptable levels of word recognition performance within 6-months post-CI activation [6].

Evaluation of proper insertion and placement of the CI electrode array is done using radiological techniques, such as computer tomography, or x-ray. The measurements of CI electrode impedance could also play a role in that. [7] Electric impedance measurement is one of the most important tools for evaluating cochlear implantation functionality [8]. EI is a measure of the resistance to the current flow in the perilymph when a voltage is applied. It is calculated as the ratio of the voltage between the stimulating and reference electrodes divided by the passed current. [9-10] the individual organic features of the inner ear may play an important role in the distribution of electrical impedance profile [11].

Impedance provides important clinical information about the device electrode function [12]. The impedances are influenced by the intracochlear position of the electrodes among other factors [13]. There is a correlation between the insertion depth with measured clinical impedances and tissue resistances. [14] Cochlear implantation using the round window (RW) route and cochleostomy achieves comparable electrode impedance and hearing results [15]. The impedance, in addition to other tests and examinations, provides the clinician with necessary information about the position of the electrode inside the cochlea. In [16], impedance measurements are used to confirm the functionality of the implanted electrodes and to give guidance for subsequent setting of the device [17]. The Robotic-Assisted Cochlear Implant Surgery (RACIS) is used in the recent years and appears to be less traumatic to intracochlear structures [18]. Electrode arrays content ranges between 12 and 22 electrode contacts. It is important to determine when the electrodes in the array are deactivated [19]. The impedance measurements help in this decision too. Reducing electrode impedance is an important factor in improving the functional benefits of CI. [20]

On average, the impedance increases between intra operative and initial session of speech processor being fitted due to protein absorption of electrode and tissue growth over electrode. then it starts deceasing gradually again with the follow up sessions until 6-12 month. Then, it stabilizes [21]. High electrode impedance values one year after implantation may imply insufficient language skill development [22]. Impedance is an indicator of overall electrode function, to detect short or open circuits, determine power consumption, guide fitting, and alert towards possible existing or impending dysfunction. [23] The impedance of a cochlear implant device can be affected by the diameter of the electrode, the position of the electrode, fibrosis surrounding the electrode and electrical stimulation [24].

Despite the importance of impedance along the CI journey and its influence on the device functionality and patients' performance. And although the advancement in the technology there is still a lack in the objective models that predict the variation of electrode impedance along the time. Therefore, the primary objective of this work is to develop a machine learning model to predict the electrode impedance after one month from the cochlear implantation surgery. The remainder of this paper is organized as follows. Section 2. Describes related works. Section 3 illustrates the Dataset. Section 4 demonstrates using Machine Learning to implement the different models. Section 5 presents experimental results and analysis. Finally, we present the future work, and conclusion in Section 6.

## II. RELATED WORK

A common usage for machine learning in cochlear implantation (CI) is signal processing. In [25] the authors did a study that includes 298 articles which are filtered to 39 articles. These articles indicate the usage of machine learning to assist CI in signal processing optimization (43.6%), automated evoked potential measurement (15.4%), postoperative performance prediction (12.8%) and surgical anatomy location prediction (7.7%) and (5.1%) in each of robotics, electrode placement performance and biomaterials performance. The most common machine learning algorithms are neural networks (47.5%), support vector machines (17.5%) and random forest/decision tree (12.5%).

It's very important to have proper emplacement of the electrode array during the Cochlear Implant (CI) surgery. The trauma-induced during the surgery could lead to residual hearing loss. In [26] The authors used machine learning to classify the insertion lengths of the electrode array into the Scala tympani. They used different machine learning algorithms like Shallow Neural Networks (SNN) and Support Vector Machines (SVM). SNN accuracy is 86.1% for partial insertion data, while SVM accuracy is 97.1% for full insertion. More than one electrode could stimulate the same area in the neural region. This interaction could have a negative impact on the hearing outcome. In [27] authors used machine learning techniques to automate Image Guided Cochlear Implant Programming (IGCIP). The IGCIP is used to help the audiologists and recommend specific CI configuration for each patient.

To predict the postoperative cochlear implant performance after one year from the surgery, authors in [28] used different machine learning algorithms like neural networks and tree-based ensemble algorithms. The accuracy of the neural network model was 95.4%. The model uses 282 preoperative variables like age at the surgery, gender, preoperative use of hearing aid, hearing loss cause, etc. The dataset includes 1604 patients who received CI from 1989 to 2019. This set is split into training dataset (85%) and test dataset (15%).

In [29] the authors used semi-supervised SVM model to predict language outcomes following CI based on pre-implant brain functional magnetic resonance (fMRI) imaging. The developed model gives 2-year prediction for developing effective language skill. The study demonstrated that semi-supervised model provides better results than the supervised model when unlabeled data are available. In [30] authors used various algorithms like SVM and random forest to predict wither healthy people will accept speech of children with cochlear implants. The evaluation is done by 80 college students who didn't know that the children did CI. The results show that random forest algorithm produce best results with RMSE = 0.108. The SVM comes after it with RMSE=0.109 while the multiple regression algorithm provides RMSE=0.113. The dataset includes 91 hearing-impaired children. The age of the children between 4 and 8 years old.

In [31] authors used machine learning model to predict cochlear implantation outcome in adults using Cross-modal cortical activity in the brain.

## III. DATASET

We have prepared a dataset of 80 patients. The studied electrode array comprises of 12 electrodes numbered from 1 to 12. It's manufactured by MED-EL (MED-EL, Innsbruck, Austria). The Dataset attributes include age at implantation and electrode impedance measured during the surgery at different channels from 1 to 12. The patients were all of the same age group, have normal cochlea, use the device for similar number of hours daily, implanted with same device, and have similar follow up duration.

Dataset preparation required many processes that have been automated using Ring programming language and using the Programming Without Coding Technology (PWCT) software which is considered as a general-purpose visual programming language [32-33]. Table 1 demonstrates the dataset features while Table 2 demonstrates the minimum and maximum values in the dataset for each feature.

TABLE I. DATASET FEATURES

| Features (GROUP 1) | Features (GROUP 2) | Label | Rows |
|---|---|---|---|
| Age at implantation | Age at implantation, EI_Intra_1, EI_Intra_2, EI_Intra_3, EI_Intra_4, EI_Intra_5, EI_Intra_6, EI_Intra_7, EI_Intra_8, EI_Intra_9, EI_Intra_10, EI_Intra_11, EI_Intra_12 | One label from EI_1M_1 to EI_1M_12 | 80 |

TABLE II. MINIMUM AND MAXIMUM VALUE FOR EACH LABEL IN THE DATASET

| Label | Minimum Value (KO) | Maximum Value (KO) | Range (KO) |
|---|---|---|---|
| EI_1M_1 | 4.48 | 16.86 | 12.38 |
| EI_1M_2 | 5.37 | 17.64 | 12.27 |
| EI_1M_3 | 4.19 | 14.57 | 10.38 |
| EI_1M_4 | 3.38 | 17.47 | 14.09 |
| EI_1M_5 | 2.71 | 16.5 | 13.79 |
| EI_1M_6 | 2.1 | 10.55 | 8.45 |
| EI_1M_7 | 1.97 | 8.94 | 6.97 |
| EI_1M_8 | 2.34 | 8.59 | 6.25 |
| EI_1M_9 | 2.34 | 8.3 | 5.96 |
| EI_1M_10 | 2.12 | 8 | 5.88 |
| EI_1M_11 | 2.12 | 9.62 | 7.5 |
| EI_1M_12 | 2.12 | 9.31 | 7.19 |

## IV. MACHINE LERNING MODELS

This study uses machine learning algorithms for regression. We have applied several ML algorithms including Linear Regression, Boosted Decision Tree regression, Decision Forest regression, Neural Network regression, and Bayesian linear regression. We used Microsoft Azure Machine Learning to create the models [34-36].

Table 3. demonstrates the results and the best algorithm to predict the impedance for each electrode channel.

TABLE III. BEST ALGORITHM AND THE RMSE

| Label | Best Algorithm | Features Group | RMSE |
|---|---|---|---|
| El_1M_1 | Bayesian Linear Regression (BLR) | 2 | 0.936972 |
| El_1M_2 | Decision Forest Regression (DFR) | 2 | 0.934948 |
| El_1M_3 | Linear Regression (LR) | 2 | 0.802687 |
| El_1M_4 | Bayesian Linear Regression (BLR) | 2 | 1.122467 |
| El_1M_5 | Bayesian Linear Regression (BLR) | 2 | 1.175844 |
| El_1M_6 | Bayesian Linear Regression (BLR) | 2 | 1.172564 |
| El_1M_7 | Bayesian Linear Regression (BLR) | 2 | 1.026447 |
| El_1M_8 | Bayesian Linear Regression (BLR) | 2 | 1.050416 |
| El_1M_9 | Bayesian Linear Regression (BLR) | 2 | 1.087652 |
| El_1M_10 | Neural Network Regression (NNR) | 2 | 0.872403 |
| El_1M_11 | Neural Network Regression (NNR) | 1 | 0.90388 |
| El_1M_12 | Boosted Decision Tree Regression (BDTR) | 2 | 0.889965 |

## V. RESULTS AND DISCUSSION

Table 4 summarizes prediction results using each of the ML algorithms. In this table we list each algorithm and the count of labels where this algorithm provides the best result compared to the other algorithms. In table 5 we list the error range at each channel and the percentage of predicting the electrode impedance in this range of errors.

These results show the best algorithm to use for predicting the electrode impedance at a specific channel. Expected error range is extracted as well. For example, to predict the electrode impedance at channel 10, we know from table 3 that the best algorithm is NNR while using group 2 of the features that includes the age at implantation and the electrode impedance at all channels from 1 to 12 which are measured during the cochlear implant surgery. The RMSE is 0.872403 KO (The range is 5.88 KO from Table 2.). Using Table 5. We know that in 58.33% of the prediction results of the electrode impedance at channel 10, the error range is between 0 and 1 KO. In 33.33% of the predicted cases the error range is between 1 and 2 KO which means that in 91.67% of the predicted cases the error range is between 0 and 2 KO. Since In 8.33% of the cases the error range is between 2 and 3 KO, this means in 100% of the cases, the error range exist between 0 and 3 KO. These results are achieved while testing the model using data that have never been seen before by the model, i.e., test dataset.

TABLE IV. BEST ALGORITHMS FOR PREDICTING THE IMPEDANCE IN THE ELECTRODE ARRAY

| Algorithm | Count of Labels |
|---|---|
| Bayesian Linear Regression (BLR) | 7 |
| Neural Network Regression (NNR) | 2 |
| Decision Forest Regression (DFR) | 1 |
| Boosted Decision Tree Regression (BDTR) | 1 |
| Linear Regression | 1 |

TABLE V. PREDICTING THE ELECTRODE IMPEDANCE AND THE ERROR PERCENTAGE IN DIFFERENT RANGES

| Label | 0-1 KO | 1-2 KO | 2-3 KO | 0-2 KO | 0-3 KO |
|---|---|---|---|---|---|
| EI_1M_1 | 20.83% | 33.33% | 20.83% | 54.16% | 75% |
| EI_1M_2 | 29.16% | 33.33% | 4.16% | 62.50% | 66.66% |
| EI_1M_3 | 41.66% | 20.83% | 20.83% | 62.50% | 83.33% |
| EI_1M_4 | 37.50% | 29.17% | 16.67% | 66.67% | 83.33% |
| EI_1M_5 | 54.17% | 20.83% | 8.33% | 75% | 83.33% |
| EI_1M_6 | 41.67% | 29.17% | 20.83% | 70.83% | 91.67% |
| EI_1M_7 | 37.50% | 29.17% | 25% | 66.67% | 91.67% |
| EI_1M_8 | 41.67% | 29.17% | 12.50% | 70.83% | 83.33% |
| EI_1M_9 | 58.33% | 20.83% | 12.50% | 79.17% | 91.67% |
| EI_1M_10 | 58.33% | 33.33% | 8.33% | 91.67% | 100% |
| EI_1M_11 | 58.33% | 33.33% | 8.33% | 91.67% | 100% |
| EI_1M_12 | 45.83% | 8.33% | 29.17% | 54.17% | 83.33% |

## VI. CONCLUSION AND FUTURE WORK

The findings of this study concluded that machine learning could be an effective and efficient tool in the cochlear implant filed. The use of machine learning models in predicting the CI impedance can help the professionals to take early decisions, objectively rather than subjectively, to improve hearing quality of CI patients.

In this paper, we presented a Machine Learning model to predict the cochlear impedance at different 12 channels using different algorithms like boosted decision tree, neural networks, and decision forest. We did many experiments to evaluate the performance of each model to determine which one provides the best results. Our results demonstrate that using a specific algorithm for prediction at each channel could provide better results. Also, the accuracy level varies between 66% and 100% based on the electrode channel when accepting an error range between 0 and 3 KO. We developed the model using the Microsoft Azure Machine Learning tool. This study can be extended in the future to provide prediction for the electrode impedance at different time frames like three months, six months and one year. Also, an application can be developed to help medicines use these models directly with new patient data.